\documentclass{llncs}

\usepackage{makeidx}
\usepackage{color}
\usepackage{amsmath}
\usepackage{amsfonts}
\usepackage{listings}
\usepackage[ruled, linesnumbered]{algorithm2e}
\usepackage{tikz}
\usepackage{pgfplots}
\usepackage{epsfig}
\usepackage{multicol}
\usepackage{multirow}
\usepackage{balance}
\usetikzlibrary{patterns}
\usepackage{wrapfig}
\usepackage{subfigure}
\pgfplotsset{every axis/.append style={
                    legend style={font=\tiny}
                    }}

\begin{document}

\title{SkILL - a Stochastic Inductive Logic Learner}

\author{Joana C\^orte-Real 
   \and Theofrastos Mantadelis
   \and In\^es Dutra
   \and Ricardo Rocha}

\institute{CRACS \& INESC TEC and Faculty of Sciences, University of Porto\\
           Rua do Campo Alegre, 1021, 4169-007 Porto, Portugal\\
           \email{\{jcr,theo.mantadelis,ines,ricroc\}@dcc.fc.up.pt}}

\maketitle


\begin{abstract}
Probabilistic Inductive Logic Programming (PILP) is a relatively
unexplored area of Statistical Relational Learning which extends
classic Inductive Logic Programming (ILP). This work introduces SkILL,
a Stochastic Inductive Logic Learner, which takes probabilistic
annotated data and produces First Order Logic theories. Data in
several domains such as medicine and bioinformatics have an inherent
degree of uncertainty, that can be used to produce models closer
to reality. SkILL can not only use this type of probabilistic data to
extract non-trivial knowledge from databases, but it also addresses
efficiency issues by introducing a novel, efficient and effective
search strategy to guide the search in PILP environments. The
capabilities of SkILL are demonstrated in three different datasets:
(i) a synthetic toy example used to validate the system, (ii) a
probabilistic adaptation of a well-known biological metabolism
application, and (iii) a real world medical dataset in the breast
cancer domain. Results show that SkILL can perform as well as a
deterministic ILP learner, while also being able to incorporate
probabilistic knowledge that would otherwise not be considered.
\end{abstract}


\section{Introduction}

Statistical Relational Learning (SRL)~\cite{Raedt-04} is a well-known
collection of techniques whose main objective is to produce
interpretable probabilistic classifiers, often in the form of readable
logical sentences. While researchers have spent their efforts on
creating logic languages to represent probabilities and runtime
environments that can deal with them~\cite{DBLP:conf/ijcai/SatoK97,Richardson-06,DBLP:conf/ijcai/RaedtKT07,DBLP:conf/ilp/KerstingR08,pfl,SLP},
few works have been dedicated to \emph{learn} rules from
probabilistic knowledge. In this work, we introduce SkILL -- a
Stochastic Inductive Logic Learner -- which can combine the rule
learning capability of classic Inductive Logic Programming
(ILP)~\cite{Lavrac-01,Muggleton-94} with uncertain knowledge as
probabilistic annotated data to produce First Order Logic (FOL)
theories.

ILP is a machine learning branch which stands out due to its
suitability to handle relational data. ILP's main goal is to
construct a \emph{theory} which explains a set of observations (called
\emph{examples}), given a set of facts and/or rules which are of a
relational nature (called \emph{background knowledge}). The induced
theory can then be used for prediction (as it can output probability
values for a given example) as well as
classification (as it can also output the specific categorical
label for an example). Probabilistic Inductive Logic Programming
(PILP)~\cite{Raedt-04} extends discrete ILP by considering
background knowledge and/or examples that are annotated with
probabilities. This is a natural extension of ILP and can in fact
model different semantic scenarios, according to the meaning that is
assigned to the probabilities.

Using probabilities to describe data has the potential advantage of
greatly reducing the dataset size, since useful information can still be
extracted from marginal distributions. Also, in cases where the full
conditional probability table is not known, information can still be
used efficiently in the computation of a rule, for instance, by
adding values from the literature in this form to the background
knowledge. Compressing data in such a way could also be used in
order to protect private sensitive data. There are surely several
other scenarios in which probabilities can be applied and 
taken advantage of. Throughout this work, probabilities will be used
as marginal distributions (motivational example), as a transformation
of a numeric attribute in discrete data (metabolism dataset), and as
an empirical confidence (non-definitive biopsies dataset).

SkILL can not only use all these types of probabilistic data but it
also addresses efficiency issues by introducing a novel, efficient and
effective search strategy to guide the search for FOL theories. SKILL
runs on top of the Yap Prolog system~\cite{santoscosta-tplp12}, uses
GILPS~\cite{Muggleton-08} as the basis rule generator and
MetaProbLog~\cite{MANTADELIS11CICLOPS,MantadelisPHD2012} (an extension
of ProbLog~\cite{DBLP:conf/ijcai/RaedtKT07,Kimmig-11}) as the
probabilistic representation language. Knowledge is thus annotated
according to ProbLog syntax and the MetaProbLog engine is used to
evaluate the probabilities of the generated theories.
  
The remainder of this paper is organized as follows. First, a toy
example is introduced to motivate the transition between ILP and PILP,
followed by a description of related work. Next, we present the SkILL
system and focus on some efficiency issues. Then, two experiments
are performed to assess SkILL's performance, followed by a discussion
of results and the conclusion.


\section{Motivational Example}
\label{sec:example}

Rock-paper-scissors is a game where two players each play one of
the three objects - either rock, paper or scissors - simultaneously,
through movements of their hands, and the winner is chosen based on the
rules presented in Fig.~\ref{fig:rps_rules} (which use the Prolog syntax).

{\small
\vspace{-0.25cm}
\begin{figure}[ht]
\begin{lstlisting}
beats(Round,PlayerA,PlayerB) :-
   plays(Round,PlayerA,rock), plays(Round,PlayerB,scissors).
beats(Round,PlayerA,PlayerB) :-
   plays(Round,PlayerA,paper), plays(Round,PlayerB,rock).
beats(Round,PlayerA,PlayerB) :-
   plays(Round,PlayerA,scissors), plays(Round,PlayerB,paper).
\end{lstlisting}
\caption{Rules of the rock-paper-scissors game in Prolog syntax}
\label{fig:rps_rules}
\end{figure}
\vspace{-0.25cm}
}

If data of this game were recorded, it would contain players' choices
of objects for each round as well as the result of each game. This is
illustrated in Fig.~\ref{fig:rps1}, where the first argument
represents each round (consecutive integers), the second argument is
the player (\lstinline+playerA+, \lstinline+playerB+ and
\lstinline+playerC+), and the third argument corresponds to each
player's outcome (\lstinline+rock+, \lstinline+paper+ or
\lstinline+scissors+). Predicate \lstinline+beats/3+ represents for
each round (first argument) which player is the winner (second
argument), and which one is the loser (third argument).

{\small
\vspace{-0.25cm}
\begin{figure}[ht]
\begin{lstlisting}
plays(1,playerA,paper).             plays(2,playerB,rock).
plays(1,playerB,scissors).          plays(2,playerC,scissors).
beats(1,playerB,playerA).           beats(2,playerB,playerC).
\end{lstlisting}
\caption{Example of full description of game}
\label{fig:rps1}
\end{figure}
\vspace{-0.25cm}
}

Traditional ILP can be used in this problem as is to induce the rules
of the game. This formulation of the problem is trivial for an ILP
engine and it can take as few as three examples to learn the three
rules of the game.

SkILL allows for inducing the same set of rules from different
background knowledge (BK) information. Suppose the information about
each round was not available and that all available information was
the profile/strategy of a given player (how often does he/she play
each object) and how often did that player win against other
players. This setting carries much less information because nothing is
known about the sequence of games or against whom a player played;
only the marginal distributions are known. Figure~\ref{fig:rps_prob}
presents an example of this new form of BK, where semi-colon has the
meaning of an exclusive-or connective (different from the Prolog syntax).

\vspace{-0.25cm}
\begin{figure}[ht]
\begin{lstlisting}
0.1::plays(playerA,rock);        0.1::plays(playerB,rock);
0.1::plays(playerA,paper);       0.3::plays(playerB,paper);
0.8::plays(playerA,scissors).    0.6::plays(playerB,scissors).

0.4::beats(playerA,playerB).
\end{lstlisting}
\caption{Probabilistic BK of rock-paper-scissors game}
\label{fig:rps_prob}
\end{figure}
\vspace{-0.25cm}

In Fig.~\ref{fig:rps_prob}, rules are annotated according to Halpern's
type 1 probability structure~\cite{Halpern-90}, where numbers on the
left correspond to values of the game domain, which can be interpreted
as the frequency with which each event happens. Predicates
\lstinline+plays+ and \lstinline+beats+ have now only 2 arguments
because the frequencies of the rounds are no longer relevant to the
problem.

Experiments were made by annotating simulated games based on random
player profiles and SkILL induced the rules presented in
Fig.~\ref{fig:rps_rules} from information about the profiles of
players using as little as 10 observations and three players.

\section{SkILL}
\label{sec:SkILL}

SkILL is a tool which can extract non-trivial knowledge (FOL theories)
from probabilistic data. As is the case of ILP systems, SkILL's
setting includes three main components:

\begin{description}
\item[Probabilistic Background Knowledge (PBK)] represents the basic
  information known about the problem and can be composed of both
  rules and facts, either probabilistic or not.
\item[Probabilistic Examples (PE)] represent the observations the
  system is attempting to explain. In the classical ILP setting there
  can be positive and negative examples, but in the probabilistic
  setting that information must be encoded as probabilities. These
  probabilities are the \emph{expected values} of examples and can
  represent either statistical information or the degree of belief in
  an example (using type I or type II probability
  structures~\cite{Halpern-90}, respectively).
\item[Search Space Constraints] mode declarations used to guide the
  search, whose aim is to minimize a loss function.
\end{description}

Since search spaces are often too large, a common approach is to guide
the search by using strategies that can lead to \emph{good hypotheses}
without exhaustively traversing all the search space. SkILL introduces
a novel, efficient and effective search strategy to guide the search
in PILP environments.


\subsection{Traversing the Search Space}
\label{sec:search_space}

Algorithm~\ref{alg:SkILL_algorithm} presents SkILL's main
algorithm. The algorithm takes as input the probabilistic background
knowledge (PBK) and a set of examples (PE) plus parameters
corresponding to the maximum length of a theory (or set of hypotheses)
to be generated (MaxTheoryLength), the number of hypotheses to be
combined in order to limit the search space (Psize and Ssize), a
metric to rank the selection of hypotheses to be combined
(RankMetric), and a final metric that is used to decide what is the
best theory found (EvalMetric).

{\small
\begin{algorithm}[ht]
\textbf{Input} = {PBK, PE, MaxTheoryLength, Psize, Ssize, RankMetric, EvalMetric}\\
\textbf{Output} = {Best theory according to EvalMetric}\\
Hyps1 = HypsN = AllHyps = generate\_hypotheses\_length\_one(PBK, PE)\\
\For{Length = 2; Length $\le$ MaxTheoryLength; Length++}{
	Primary = select\_primary\_set(HypsN, Psize, RankMetric)\\
	Secondary = select\_secondary\_set(Hyps1, Ssize, RankMetric)\\
	HypsN = generate\_combinations(Primary, Secondary)\\
	AllHyps = AllHyps $\cup$ HypsN\\
}
\textbf{return} best\_theory(AllHyps, EvalMetric)\\
\caption{SkILL Algorithm}
\label{alg:SkILL_algorithm}
\end{algorithm}
}

Initially, the algorithm uses the TopLog engine from the
GILPS~\cite{Muggleton-08} ILP system, to generate all possible
hypotheses composed of only one clause (line 3 in
Alg.~\ref{alg:SkILL_algorithm}). A top level generic hypothesis is
constructed from the mode declarations in the PBK and possible
hypotheses are generated independently from each example using SLD
refutation. This approach ensures that each hypothesis generated must
be entailed by at least one example, and so the hypotheses mirror
patterns contained in the observations with respect to the PBK. SkILL
improves on this approach by removing hypotheses which are
permutations of each other (i.e., syntactically distinct but
semantically equal), so that probabilistic inference is only performed
over semantically unique hypotheses.

Once hypotheses with length one are generated, the algorithm proceeds
by generating hypotheses with length greater than one (lines 4--9 in
Alg.~\ref{alg:SkILL_algorithm}) until reaching a given maximum theory
size (argument MaxTheoryLength). Combining hypotheses in order to
generate new hypotheses with larger size is not a trivial task --
possible combinations are $N \choose K$ with $N$ being the total
number of length one hypotheses and $K$ the maximum theory
size. Ideally, an exhaustive search of the hypotheses space would be
performed, but this is computationally taxing, particularly as the
theory size grows. Therefore, SkILL's search strategy selects
candidate hypotheses for two different sets, named \emph{Primary} and
\emph{Secondary}, and new hypotheses are then generated by only
combining members of these sets. To do so, for each theory length, the
algorithm first selects the Primary and Secondary sets of hypotheses,
with sizes equal to arguments Psize and Ssize, respectively (lines
5--6 in Alg.~\ref{alg:SkILL_algorithm}), and then it performs the
combinations (line 7 in Alg.~\ref{alg:SkILL_algorithm}). This
procedure repeats until generating hypotheses for all lengths.

SkILL's selection procedure has two main goals: (i) reduce the number
of combinations to be generated without losing the \emph{good
  hypotheses} in the process and (ii) introduce some stochastic
behavior by giving identical opportunity to \emph{weaker rules} whose
combination can be of interest. The primary and secondary sets can be
seen as a way to materialize these two goals, respectively.

The primary set of hypotheses is considered to be the most relevant,
i.e., the one holding the \emph{best set of hypotheses} according to a
given \emph{ranking metric} (argument RankMetric). In each iteration
of the algorithm, the primary set is filled with the Psize best
hypotheses from the set of hypotheses generated in the previous
iteration (1 clause hypotheses when searching for 2 clauses
hypotheses; 2 clauses hypotheses when searching for 3 clauses
hypotheses; etc). To rank hypotheses, SkILL supports three metrics:
RMSE (root mean square error), PAcc (probabilistic accuracy) and
Random.

The secondary set is filled with Ssize hypotheses from the set of
hypotheses with length one. The aim of the secondary set is to include
very different candidate hypotheses whose combination with the
hypotheses from the primary set can be of interest. Priority to full
stochastic behaviour can be given by randomly selecting all the
hypotheses for the secondary set, or a selection based on best set of
hypotheses according to the given ranking metric can be
made. Additionally, both approaches can be combined in order to obtain
a more heterogeneous set.

In particular, the experimental results presented used a mixed
scenario where the secondary set always includes the Psize best
hypotheses with one clause (i.e., the hypotheses selected for the
first primary set) plus (Ssize - Psize) randomly selected distinct
candidates from the remaining hypotheses with one clause. This
stochastic component of the selection is distinct for each iteration.

Finally, according to a given evaluation metric (argument EvalMetric),
the best generated hypothesis for all different lengths is returned
(line 10 in Alg.~\ref{alg:SkILL_algorithm}). 


\subsection{Evaluation Metrics}
\label{sec:metrics}

Currently, SkILL implements the RMSE and PAcc metrics. These metrics
can be used to rank and/or evaluate hypotheses, as mentioned
earlier. Since, from the point of view of SkILL's algorithm, the
ranking and evaluation phases are independent, we have chosen to
introduce two different metric arguments instead of only one. By doing
this, we not only highlight that independence but also do not restrict
possibly different metric combinations.

The RMSE metric penalizes predictions farther from the expected
values, while PAcc is the generalization of the discrete accuracy to
the probabilistic setting as introduced by De~Raedt and
Thon~\cite{Raedt-11} and used by Muggleton~\cite{Muggleton-14a}.

The RMSE of a hypothesis $H$ can be defined as:
{\small
\begin{equation} \label{eq:rmse_prior}
\begin{split}
RMSE_H & = \frac{1}{|PE|} \sum_{e_i \in PE} (P_{H}(e_i) - P(e_i))^2\\
\end{split}
\end{equation}
}
where, $P_{H}(e_i)$ denotes the probability that $H$ together
with the $PBK$ entails an example $e_i$, and $P(e_i)$ denotes the
given expected value of an example $e_i$.

The PAcc of a hypothesis $H$ is often represented in terms of true
positive ($TP$), true negative ($TN$), false positive ($FP$) and false
negative ($FN$) examples, as shown in Equation~\ref{eq:pacc_proof1}.
{\small
\begin{equation} \label{eq:pacc_proof1}
\begin{split}
PAcc_H & = \frac{TP + TN}{TP + TN + FP + FN}\\
\end{split}
\end{equation}
}
From~\cite{Raedt-11}, $TP + TN + FP + FN = |PE|$, and $TP$ and $TN$
are equal to the sum over all examples of $min(P_{H}(e_i),
P(e_i))$ and $min(1-P_{H}(e_i), 1-P(e_i))$,
respectively. Substituting into Equation~\ref{eq:pacc_proof1}, this
gives that PAcc can be also represented in terms of the
absolute average error between predictions and expected values, as
shown in Equation~\ref{eq:pacc_proof2}.
{\small
\begin{equation} \label{eq:pacc_proof2}
\begin{split}
PAcc_H = & \frac{1}{|PE|} \sum_{e_i \in PE} ( min(P_{H}(e_i), P(e_i)) + min(1-P_{H}(e_i), 1-P(e_i)) ) \\
= & \frac{1}{|PE|} \sum_{e_i \in PE} (min(P_{H}(e_i), P(e_i)) + 1-max(P_{H}(e_i), P(e_i))\\
= & \frac{1}{|PE|} \sum_{e_i \in PE} (1 - |P_{H}(e_i) - P(e_i)|)\\
= & 1 - \frac{1}{|PE|} \sum_{e_i \in PE} |P_{H}(e_i) - P(e_i)|
\end{split}
\end{equation}
}
As presented in Equation \ref{eq:rmse_pacc}, both metrics can also be
defined based on a common loss function $loss_H (e_i ) = P_{H}(e_i) -
P(e_i)$, which calculates the difference between the probabilistic
expected value of an example and the value that can be predicted w.r.t
a given hypothesis and the PBK.
{\small 
\begin{equation} \label{eq:rmse_pacc}
\begin{split}
RMSE_H = \frac{1}{|PE|} \sum_{e_i \in PE} loss_H(e_i)^2\\
PAcc_H = 1- \frac{1}{|PE|} \sum_{e_i \in PE} |loss_H (e_i)|
\end{split}
\end{equation}
}
Hence, the aim of SkILL's search engine is to find the hypothesis with
minimum RMSE or maximum PAcc in the search space.


\subsection{Pruning Combinations}
\label{sec:pruning}

PILP shares a similar hypothesis search space as an ILP problem where
all examples are expected to be true. The difference between the
approaches lies obviously in the evaluation of hypotheses; in the
first case it is a number between $0.0$ and $1.0$ representing a
probability, whilst in the latter it is either true or false.

Theories in ILP are constructed by combining several hypotheses
through logic conjunction ($\wedge$) and disjunction ($\vee$). Let
$H_1, H_2$ be hypotheses; the hypothesis resulting from conjuncting
$H_1$ with $H_2$ is more specific than either $H_1$ or $H_2$, while
the disjunction of $H_1$ and $H_2$ is more general than either $H_1$
or $H_2$. Equation~\ref{eq:disc_hyps} shows how an example $e_i$ could
be entailed by a disjunction in terms of two hypotheses $H_1$ and
$H_2$.
{\small
\begin{equation} \label{eq:disc_hyps}
\begin{split}
H_1 \vee H_2 \models e_i \Rightarrow & H_1 \wedge \bar{H_2} \models e_i \hspace{4pt} OR\\
&\bar{H_1} \wedge H_2 \models e_i \hspace{4pt} OR\\
&H_1 \wedge H_2 \models e_i 
\end{split}
\end{equation}
}
Equation~\ref{eq:disc_hyps} can be extended to the probabilistic case
using the principle of inclusion/exclusion as follows.  In the
probabilistic scenario, the probability of a hypothesis $P_H$
represents the probabilistic mass covered by $PBK \cup H \models
true$, which can range between 0 and 1. As such, the probability of a
disjunction of hypotheses can be calculated according to the
expression shown in Equation~\ref{eq:inc_exc}.
{\small
\begin{equation} \label{eq:inc_exc}
P_{H_1 \vee H_2}(e_i) = P_{H_1}(e_1) + P_{H_2}(e_i) - P_{H_1 \wedge H_2}(e_i)
\end{equation} 
}
The $P_{H_1 \wedge H_2}(e_i)$ term of Equation~\ref{eq:inc_exc} is the
probability of both hypotheses (conjunction) entail the example. From
set theory, three particular cases are known, namely: 
completely overlapping, independent or disjoint masses, and these can be
calculated according to the expressions in
Table~\ref{tab:prob_expressions}.
{\small
\begin{table}
\centering
\caption{Special cases of $P_{H_1 \vee H_2}(e_i)$ from set theory}
\label{tab:prob_expressions}
\begin{tabular}{c|c|c}
  Completely Overlapping & Independent & Disjoint \\ \hline
  $max(P_{H_1}(e_i), P_{H_2}(e_i))$ & $P_{H_1}(e_i) + (1-P_{H_1}(e_i))(P_{H_2}(e_i))$ & $P_{H_1}(e_i) + P_{H_2}(e_i)$\\
\end{tabular}
\end{table}
\vspace{-0.25cm}
}
By analysing Table~\ref{tab:prob_expressions}, it becomes evident that
the probability of the disjunction of two hypotheses has clear minimum
and maximum boundaries, derived from the cases of completely
overlapping and disjoint masses, respectively.
{\small
\begin{equation} \label{eq:boundaries}
P_{H_1 \vee H_2}(e_i) \in [max(P_{H_1}(e_i), P_{H_2}(e_i)), min(P_{H_1}(e_i) + P_{H_2}(e_i), 1.0)]
\end{equation}
}
The boundaries stated in Equation~\ref{eq:boundaries} make it possible
to prune combinations of hypotheses whose interval of results does not
contain the expected value for that example. SkILL can use these
boundaries to prune combinations in two different contexts: (a)
\emph{before probabilistic inference}, to avoid performing such
computations on some combinations, and (b) \emph{after probabilistic
  inference}, to remove combinations found to be \emph{bad} after
inference. This is part of the \emph{generate\_combinations()}
function as illustrated in
Algorithm~\ref{alg:SkILL_pruning}. Functions
\emph{possibly\_good\_combination()} and \emph{good\_combinations()}
at lines 8 and 15 in Alg.~\ref{alg:SkILL_pruning} implement each
pruning strategy, respectively.
{\small
\begin{algorithm}[ht]
\textbf{Input} = {Primary and Secondary sets}\\
\textbf{Output} = {Combination of (not pruned) hypotheses from both sets}\\
\Begin{
	HypsN = \{\}\\
	\ForEach{$H_p$ in Primary} {
		\ForEach{$H_s$ in Secondary} {
		  	$H_{new} = H_p \vee H_s$\\
			\If{possibly\_good\_combination($H_{new}$)}{
				$H_{(new,prob)} =$ do\_problog\_inference($H_{new}$)\\
				HypsN = HypsN $\cup$ $H_{(new,prob)}$
			}
		}
	}
}
\textbf{return} good\_combinations(HypsN)\\
\caption{Function \emph{generate\_combinations()}}
\label{alg:SkILL_pruning}
\end{algorithm}
}

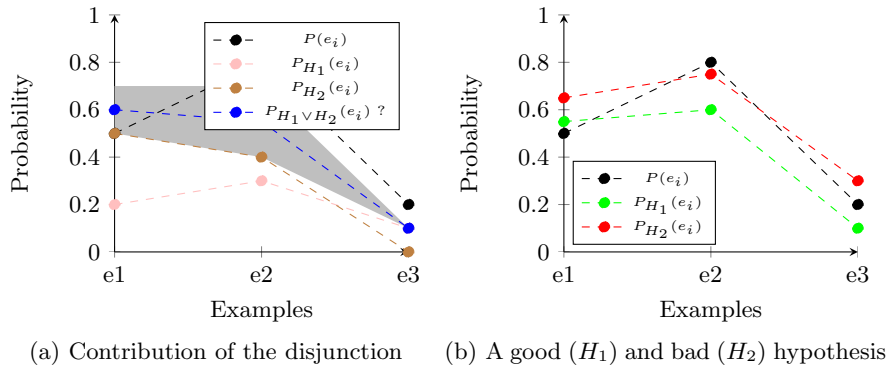
\begin{figure*}
\centering
\subfigure[Contribution of the disjunction]{  \label{fig:pruning_a}
  \centering
    \begin{tikzpicture}
	\begin{axis}[
			legend style={
			legend pos=north east,
			},
			width=0.45*\textwidth,
			axis x line=bottom,
			axis y line=left,
			ymax = 1,
			ymin = 0,
			symbolic x coords = {e1, e2, e3},
			xlabel=Examples,
			ylabel=Probability,
			xtick=data
			]
			\addplot[fill=lightgray,draw=none] coordinates {(e1, 0.5) (e1,0.7) (e2, 0.7)  (e3, 0.1) (e2, 0.4) (e1, 0.5)} \closedcycle;
			\addplot [dashed, mark=*, color=black] plot coordinates {(e1, 0.5) (e2, 0.8) (e3, 0.2)};
			\addplot [dashed, mark=*, color=pink] plot coordinates {(e1, 0.2) (e2, 0.3) (e3, 0.1)};
			\addplot [dashed, mark=*, color=brown] plot coordinates {(e1, 0.5) (e2, 0.4) (e3, 0)};

			\addplot [dashed, mark=*, color=blue] plot coordinates {(e1, 0.6) (e2, 0.55) (e3, 0.1)};
			\legend{, $P(e_i)$, $P_{H_1}(e_i)$, $P_{H_2}(e_i)$,  $P_{H_1 \vee H_2}(e_i)$ ?}
		\end{axis}
	\end{tikzpicture}
}
\subfigure[A good ($H_1$) and bad ($H_2$) hypothesis]{ \label{fig:pruning_b}
  \centering
  \begin{tikzpicture}
	\begin{axis}[
			legend style={
			legend pos=south west,
			},
			width=0.45*\textwidth,
			axis x line=bottom,
			axis y line=left,
			ymax = 1,
			ymin = 0,
			symbolic x coords = {e1, e2, e3},
			xlabel=Examples,
			ylabel=Probability,
			xtick=data
			]
			\addplot [dashed, mark=*, color=black] plot coordinates {(e1, 0.5) (e2, 0.8) (e3, 0.2)};
			\addlegendentry{ $P(e_i)$}
			\addplot [dashed, mark=*, color=green] plot coordinates {(e1, 0.55) (e2, 0.6) (e3, 0.1)};
			\addlegendentry{ $P_{H_1}(e_i)$}
			\addplot [dashed, mark=*, color=red] plot coordinates {(e1, 0.65) (e2, 0.75) (e3, 0.3)};
			\addlegendentry{ $P_{H_2}(e_i)$}
		\end{axis}
	\end{tikzpicture}
}
\caption{Using the boundaries}
\label{fig:pruning}
\end{figure*}

Finding the best pruning strategies to use these boundaries is not
evident, since we must take into account the predictions of a
hypothesis for \emph{all} examples. Because data will most likely not
be completely independent or completely mutually exclusive, the
strategies must consider that the contribution of a hypothesis in a
combination of two hypotheses varies greatly and so care must be taken
not to prune away rules which might have been important. This concept
is better illustrated in Fig.~\ref{fig:pruning}.

Figure~\ref{fig:pruning_a} shows the case of combination of
hypotheses, where the shaded area represents the possible contribution
of the disjunction of hypotheses and the blue points are the estimated
values $P_{H_1 \vee H_2}(e_i)$ for the disjunction, which in the case
of our function \emph{possibly\_good\_combination()} are the values in
the center of that interval.

Since hypotheses are being combined using disjunctions, the value of
the combination for one particular example $e_i$ can only be greater
or equal than the value of any of the hypotheses in the
combination. As such, combinations of hypotheses whose result is lower
than the expected values are in principle of greater interest for
combination than others. Figure~\ref{fig:pruning_b} shows the case of
a good and a bad hypothesis according to this principle. SkILL's
pruning functions \emph{possibly\_good\_combination()} and
\emph{good\_combinations()} reflect this by discarding the
combinations $H_{new}$ whose estimated contribution is overall less than the
expected values $P(e_i)$, as shown by Equation~\ref{eq:pruning}.
{\small
\begin{equation} \label{eq:pruning}
\begin{split}
\sum_{e_i \in PE} &P_{H_{new}}(e_i) - P(e_i) > 0
\end{split}
\end{equation}
}


\section{Experimental Settings}
The foremost focus of SkILL is the discovery of non-trivial knowledge from a dataset in order to explain observations. The quality of the knowledge discovered is currently evaluated by two different metrics (probabilistic accuracy or RMSE).

Furthermore, the FOL theories found by SkILL can also be used in classification by introducing a threshold. The threshold could be learned from the original observations or be arbitrary chosen. This approach has a benefit over classical ILP (such as Aleph) in its capability to cope with noise in the data.

As such, this work presents experiments of both types: the metabolism dataset is used to evaluate the classification accuracy of the system, and a medical dataset of non-definite biopsies is used as the basis for extraction of non-trivial knowledge in this domain.

Accuracy is used to evaluate the classifiers, using the standard formula in the discrete case (Aleph) and its probabilistic extension as presented in Section~\ref{sec:SkILL} for the probabilistic case (SkILL).

\subsection{Classification}
The dataset used to assess SkILL's classification accuracy is the metabolism dataset, and is taken from the 2001 KDD Cup Challenge\footnote{\url{http://www.cs.wisc.edu/$\sim$dpage/kddcup2001}}. Although the challenge involved learning 14 different protein functions, this experiment focuses on a subtask that is to predict which proteins are responsible for metabolism. For this purpose, we use a subset of the full dataset containing 230 examples split evenly between positives and negatives. Since the dataset is originally discrete, a normalization to the \lstinline+interaction(gene1, gene2, type, strength)+ fact in the BK was made: \lstinline+interaction+'s fourth argument is a numerical argument which represents the strength of the
interaction between two genes. By transforming \lstinline+interaction(gene1, gene2, type, strength)+ to \lstinline+strength_norm::interaction(gene1, gene2, type)+, not only the search space of hypotheses is reduced (because that feature is no longer directly considered in the hypotheses generation process), but
also predicates used to typically compare numerical features in ILP
are made redundant in this case, since SkILL implicitly attempts to
find the hypotheses with the best fit to the examples, taking into
account the probabilities of the facts in the PBK. Finally, we converted the examples from discrete true/false to probabilistic with 1.0/0.0 probabilities respectively.

Metabolism is a fairly small dataset: it is composed of 230 examples
(half positive and half negative) and approximately 7000 BK facts, of
which \~3200 are probabilistic. As such, 30 70-30 bootstraps were
generated and the results presented for all experiments are the
average and standard deviation over the 30 bootstraps test sets (70\%
of each booststrap cases were used for training and 30\% for
test). Since SkILL provides several configuration options, various
scenarios were tested in order to compare the results among
them. Results presented for the Aleph system~\cite{Aleph}, are collected with the default parameters (except noise, which is set to maximum).
However, since the BK of
metabolism has been altered, the systems are not working with
comparable data, and so these values are meant to be merely
informative.

Table~\ref{tab:pruning} presents a comparison between using a pruning
strategy and exhaustively combining Primary (20 hypotheses) against
Secondary (200 hypotheses), for hypotheses until size 3. The number of
hypotheses of size 1 of the training sets ranges from 2000 to 3000,
and so Secondary represents about 10\% of hypotheses, while Primary
represents 1\%. This table also presents discrete ILP results using
Aleph's default configuration and allowing for maximum noise.

\vspace{-0.25cm}
\begin{table}[h]
\centering
\caption{Accuracy of the models on the test set for the metabolism dataset}
\label{tab:pruning}
\begin{tabular}{c|ccc}
        & \multicolumn{3}{c}{Search Strategies}\\
        & (RMSE, $\sigma$) & & (PAcc, $\sigma$)\\ \hline\hline
SkILL & (0.616, 0.063) &  & (0.661, 0.045)  \\
SkILL+pruning & (0.581, 0.099)  & & \textbf{(0.663, 0.045)}  \\\hline
Aleph & \multicolumn{3}{c}{(0.656, 0.047)}\\
\end{tabular}
\end{table}
\vspace{-0.25cm}

The results in Table~\ref{tab:pruning} show that driving the search with PAcc metric (both for evaluation and ranking) produces better classification results (2-tailed t-test, p = 0.04) than both the discrete case and when using RMSE as a ranking metric. We believe that penalizing greater distances from the expected values (like when using RMSE) produces worse results accuracy-wise because of overfitting the training dataset.

Table~\ref{tab:population_sizes} studies the effect of varying the sizes of Primary and Secondary, and how their accuracy and RMSE relate to the sizes of these sets for different ranking metrics.

\vspace{-0.25cm}
\begin{table}[h]
\centering
\caption{Probabilistic Accuracy of the test set for varying sizes of Primary and Secondary sets and different search strategies using SkILL with pruning}
\label{tab:population_sizes}
\begin{tabular}{c|cccc}
    & \multicolumn{4}{c}{Search Strategies}\\
Psize/Ssize  & (RMSE-Rand, $\sigma$) & (RMSE-RMSE, $\sigma$) & (PAcc-Rand, $\sigma$) & (PAcc-PAcc, $\sigma$) \\ \hline\hline
10/100 & (0.586, 0.093) & (0.583, 0.095) & (0.663, 0.045) & (0.663, 0.045) \\
20/200 & (0.583, 0.104)       & (0.581, 0.099)       & (0.663, 0.045)         & (0.663, 0.045)        \\
30/300 & (0.575, 0.096)       & (0.612, 0.065)       & (0.663, 0.045)         & (0.663, 0.045)       
\end{tabular}
\end{table}
\vspace{-0.25cm}

From Table~\ref{tab:population_sizes}, it becomes evident that all PAcc measurements are the same - this is because the best classifier for this dataset is always a hypothesis of length one, and therefore is always considered independently of the population and the ranking metric. None of the candidate hypotheses of length greater than one results in a better accuracy for this evaluation metric. However, when using the RMSE evaluation metric, many different hypotheses are generated, for different training datasets. Again, this substantiates the notion that the RMSE evaluation metric may be causing overfitting. These results also indicate that the difference between using a random or RMSE ranking criterion is negligible for small populations sizes (2-tailed t-test with p=0.47 and p=0.89 for 10/100 and 20/200, respectively). This happens because the best hypotheses ranked by RMSE are not good candidates for combination in this case, so the random hypotheses are in fact being used in most cases. In the case of 30/300 population, the RMSE ranking shows an improvement in the results, but at the cost of longer runtime. A random ranking strategy does not require that the population be ordered, and when the size of generated hypotheses grows, so does the time spent in ordering them.

\subsection{Knowledge extraction}

Breast cancer diagnosis guidelines suggest that patients presenting suspicious breast lesions should be sent to perform a diagnostic mammogram and possibly an ultrasound, and a core needle biopsy to further define this abnormality. The biopsy is very important in determining malignancy of a lesion and usually yields definitive results; however, in 5\% to 15\% of cases, the results are non-definitive~\cite{Poole-14}. Routine practice usually sends all patients with non-definitive biopsies to excision, even though only a small fraction of them (10-20\%) have in fact a malignant finding confirmed after the procedure - the remainder of them did not need to be subjected to surgery. In the US this represents approximately 35,000 to 105,000 women who likely underwent excision and a majority of them ultimately received a benign diagnosis. 

Although non-definitive biopsies are relatively rare, sending every woman that has a non-definitive biopsy to excision is not a good practice. Machine learning methods have been used to mitigate this and other problems by allowing to produce models of the data that can distinguish between benign and malignant
cases~\cite{Kuusisti-13,Ferreira-12}. However, in the medical domain it is crucial to represent data in a way that experts can understand and reason about, and as such ILP can successfully be used to produce such models. Furthermore, probabilistic ILP allows for incorporating in the PBK the confidence of physicians in observations and known values from the literature.

In this study, we use 130 biopsies dating from January 2006 to December 2011, which were prospectively given a non-definitive diagnosis at radiologic-histologic correlation conferences. 21 cases were determined to be malignant after surgery, and the remaining 109 proved to be benign. For all of these cases, several sources of variables were systematically collected including variables related to demographic and historical patient information (age, personal history, family history etc), mammographic BI-RADS descriptors (mass shape, mass margins, calcifications etc), pathological information after biopsy
(type of disease, if it is incidental or not, number of foci etc),
biopsy procedure information (needle gauge, type of procedure etc), and
other relevant facts about the patient. Probabilistic data was also
gathered: namely the confidence in malignancy for each case (before
excision), assigned by different physicians analysing that
case. Furthermore, and since physicians base their conclusions in
literature values from the universe of all biopsies, values were added
in the PBK as the probability of malignancy given a feature value
(\lstinline+is_malignant+ features). For example, it is well known
among radiologists expert in mammography that if a mass has a spiculated margin, the probability that the associated finding is malignant is around 90\%.

Two kinds of experiments were performed on this dataset: (i) the
\lstinline+malignancy+ experiment consisted of finding theories by
using as examples a discrete class variable malignancy determined
after excision (either malignant or not), and (ii)
\lstinline+malignancyPH+ experiments using as examples the
probabilities assigned by different physicians (PH1, PH2, PH3) to the
malignancy of each case. The resulting theories are presented in
Figure~\ref{fig:med_full}. These experiments were performed on the full
training set, since they were intended to be exploratory. For each classifier, we report accuracy on
the full training set only to illustrate differences between the
different classifiers.
{\small
\begin{figure}[ht]
\begin{lstlisting}
malignancy(Patient) :-
	distrib_Grp(Patient,missing),
	aDH(Patient,'Y'),
	domiFinding(Patient,mass).

malignancyPH1(Patient):-
	is_malignant_oval(present).
malignancyPH2(Patient):- 
	fibro(Patient,'N'),
	is_malignant_oval(present).
malignancyPH3(Patient):-
	shape_Irr(Patient,present),
	is_malignant_irregular(present).
\end{lstlisting}
\caption{Hypotheses for malignancy of non-definitive biopsies}
\label{fig:med_full}
\end{figure}
}
Figure~\ref{fig:med_full} shows the best hypotheses found using: PAcc metric both for ranking and evaluation; primary/secondary population of 20/200; and generating hypotheses until length 3.
The \lstinline+malignancy+ predicate is the best classifier found for
malignancy of a tumour (experiment (i), accuracy =
88\%). Probabilistic BK did not play an important role in this task,
since the class variable is deterministic. Nevertheless, SkILL managed to find a good rule that combines a variable/value indicative of malignancy: the presence of atypical ductal hyperplasia (\lstinline+aDH+), with neutral variables such as the presence of mass or calcification distribution grouped (\lstinline+distrib_Grp+).

Predicates \lstinline+malignancyPH+ are the classifiers found for experiment (ii) (accuracies of 94\%, 95\% and 86\%, respectively). In all \lstinline+malignancyPH+ rules, at least one of the probabilistic literals is present. For example, the probabilistic literal that corresponds to a tumour of oval shape, which is highly correlated with malignancy, appears in \lstinline+malignancyPH1+ and \lstinline+malignancyPH2+. The same happens to the literal that represents the probability of malignancy of a tumour having an
irregular shape, which appears in \lstinline+malignancyPH3+. These results express the different mental models associated with each physician. These rules seem to indicate that some physicians give more weight to shape irregular while others give more importance to shape oval, besides giving weight to the Fibroepithelial lesions \lstinline+fibro+. One of the great outcomes of these rules is that they can be combined and perhaps produce an even better model for all physicians.


While it was not evident, the system found that a length 1 theory
was sufficient to describe best the datasets studied, which is very
important, specially in the case of the medical dataset, since
physicians need to spend less time sieving through smaller
rules. However, it is obvious that there exist problems that would
require classifiers with multiple rules such as the motivational
example of Section~\ref{sec:example}. In this aspect, SkILL takes
advantage of its clever search and pruning of hypotheses combinations,
being able to explore a more qualitative portion of the full space, whilst being
able to perform both classification and prediction, efficiently
extending the classical ILP approach. 


\section{Related Work}

The PILP setting was first introduced in \cite{Raedt-04}, where three
distinct settings -- extended from traditional ILP~\cite{Lavrac-01} --
are put forward: \emph{probabilistic entailment}, \emph{probabilistic
  interpretations}, and \emph{probabilistic proofs}.  Later, Raedt and
Thon presented the system ProbFOIL~\cite{Raedt-11}, which is not only
capable of performing induction over probabilistic examples, but also
on background knowledge encoded as ProbLog probabilistic facts. A
number of relevant metrics such as precision, accuracy and m-estimate
are adapted from the discrete ILP domain for use in the new setting,
and ProbFOIL's search for a hypothesis is guided based on
probabilistic accuracy of the theories. This system then presents a
proof of concept by analyzing two toy examples and extracting First
Order Logic (FOL) rules about them. However, this system does not take
advantage of the probabilistic data in order to tune its search
engine, using simply an extension of an ILP algorithm with a different
loss function.

Probabilistic Explanation Based Learning (PEBL)~\cite{Kimmig-07} can
find the most likely FOL clause which explains a set of positive
examples in terms of a database of probabilistic facts. The
explanation clause is the combination of predicates which yields the
highest probability based on the examples, and is found by
constructing variabilized refutation proofs for the given examples
using SLD resolution. However, since PEBL is a deductive system,
information about the expected structure of the explanation should be
provided as predicates (which are often recursive).

Orthogonally, Markov Logic Networks (MLNs)~\cite{Richardson-06} also
combine structure learning using a FOL framework with a probabilistic
Markov Random Fields approach~\cite{Kok-05}. An MLN is a set of pairs
of logic formulae and weights, where the latter are calculated based
on the number of true groundings of the respective formula. Pairs
sharing at least one variable in the same grounding are connected by
an edge, and in fact an MLN can be thought of as generating a grounded
Markov Network for each possible set of facts. Structure learning of
MLNs can be done by altering the logic search space by (i) adding or
removing one or more literals from a logic formula in a pair and (ii)
inverting predicate symbols of a formula; both techniques are similar
to operations performed in traditional ILP structure
learning. Structure learning for MLNs softens the hypotheses by using
probabilities and as such produces better classifiers, as shown
in~\cite{Kok-05}; however, MLNs still consider crisp background
knowledge, not taking into account the possibility of probabilistic
logic facts. Additionally, and whilst MLNs are capable of structure
learning, the final classifier is an MLN itself, which does not have
the advantage of readability, especially when problem sizes are
larger.

Finally, Meta-Interpretive Learning~\cite{Muggleton-14a} -- which is a
technique aimed at performing predicate invention in ILP using
abduction -- can also be used to perform probabilistic structure
learning by calculating prior and posterior distributions on the
hypotheses space according to the examples explained by a given
hypothesis~\cite{Muggleton-14b}. Meta-Interpretive Learning makes it
possible to use Bayesian theory to both sample the hypotheses space
and evaluate hypotheses according to their coverage of
examples. Hypotheses search space can then be summarized as a
super-imposed logic program, where the arcs connecting atoms contain
the sum of all arcs for each individual hypothesis. As such, this
approach can learn simultaneously the structure of the arguments of
the meta rules and the parameters of the super-imposed logic
program. This approach is similar to structure learning for MLNs in
the sense that a relation exists between simultaneously grounded
entities in the data and that hypotheses are ranked according to how
many of these possible configurations they explain. However,
probabilistic background knowledge is also not supported by
meta-interpretive learning, since it does not support probabilistic
facts.


\section{Conclusions}

This work presented the PILP learner SkILL, which extends classic ILP learners by incorporating probabilistic facts and rules in its BK, as well as by using probabilistic examples.
There are different semantics which can apply to probabilistic data, and a toy example of probabilities used as marginal distributions was presented to motivate the use of data annotated with probabilities.
Then, some details on the setting SkILL uses were presented, namely focusing on the strategy to traverse the search space, the evaluation metrics applied to hypotheses and how to efficiently prune the search.
SkILL generates theories by combining hypotheses using a ranking metric and always maintaining a number of random hypotheses so as to ensure that weaker candidates are still considered. 
The evaluation metrics used to select the best hypotheses and to guide the search are probabilistic accuracy (PAcc) and root mean square error (RMSE); they differ because RMSE penalizes more heavily greater errors.
Since SkILL works on probabilistic data, a pruning strategy based on set theory and the principle of inclusion/exclusion was devised and implemented with the aim of increasing efficiency in the system.
SkILL's classification performance was assessed using a subset of the metabolism dataset after some data were adapted to probabilities. Experiments show that SkILL's accuracy performance is better than that of the discrete ILP system Aleph, and that the pruning strategy does not significantly alter SkILL's final results.
Finally, SkILL was used to extract non-trivial knowledge from a
dataset of non-definitive biopsies annotated with probabilistic
literature values. 
 Results show that rules generated from data annotated with physicians degrees of
 belief vary, but agree with medical literature values. We have been
 working on the validation of these rules on a new unseen biopsy dataset.

\bibliographystyle{plain} 
\bibliography{references}

\end{document}